\pdfoutput=1

\documentclass[11pt]{article}

\usepackage{acl}

\usepackage{times}
\usepackage{latexsym}
\usepackage[T1]{fontenc}

\usepackage[utf8]{inputenc}

\usepackage{microtype}

\usepackage{inconsolata}
\usepackage{booktabs}
\usepackage{amsfonts}
\usepackage{graphicx} 
\usepackage{soul}
\usepackage{xcolor}
\graphicspath{{figures/}}
\usepackage{array}
\usepackage{tabulary}
\usepackage{subfigure}

\newcommand{\truepred}[1]{
  \begingroup
  \sethlcolor{green}
  \hl{#1}
  \endgroup
}
\newcommand{\falsepred}[1]{
  \begingroup
  \sethlcolor{pink}
  \hl{#1}%
  \endgroup
}

\title{MedDec: A Dataset for Extracting Medical Decisions \\from Discharge Summaries}

\author{
Mohamed Elgaar\textsuperscript{$\dagger$} \;\;\;  
Jiali Cheng\textsuperscript{$\dagger$} \;\;\;  
Nidhi Vakil\textsuperscript{$\dagger$}\\
\textbf{Hadi Amiri}\textsuperscript{$\dagger$} \;\;\;  
\textbf{Leo Anthony Celi}\textsuperscript{$\ddagger$}\\ 
\textsuperscript{$\dagger$}Miner School of Computer \& Information Sciences, University of Massachusetts Lowell \\
\textsuperscript{$\ddagger$}Institute for Medical Engineering and Science, Massachusetts Institute of Technology \\
   \texttt{\{melgaar,jcheng2,nvakil,hadi\}@cs.uml.edu} \; \texttt{lceli@mit.edu}\\
  }

\begin{document}
\maketitle
\begin{abstract}
Medical decisions directly impact individuals' health and well-being. Extracting decision spans from clinical notes plays a crucial role in understanding medical decision-making processes. In this paper, we develop a new dataset called ``MedDec,'' which contains clinical notes of eleven different phenotypes (diseases) annotated by ten types of medical decisions. We introduce the task of medical decision extraction, aiming to jointly extract and classify different types of medical decisions within clinical notes. We provide a comprehensive analysis of the dataset, develop a span detection model as a baseline for this task, 
evaluate recent span detection approaches, and employ a few metrics to measure the complexity of data samples. Our findings shed light on the complexities inherent in clinical decision extraction and enable future work in this area of research. The dataset and code are available through \url{https://github.com/CLU-UML/MedDec}.
\end{abstract}

\section{Introduction}
Clinical notes 
contain rich information about medical decision-making. Such notes document patient conditions, medications, laboratory and diagnostic results, assessments and plans, prognoses, and follow-up information, among other crucial data points. However, automatic knowledge extraction from clinical notes has been challenged by imprecise clinical descriptions, heterogeneous data, and the need for data annotation. 
In particular, although there exist comprehensive and manually verified taxonomies of medical decisions~\citep{braddock1997doctors,ofstad2018clinical}, and successful information extraction techniques in medical~\citep{mullenbach-etal-2021-clip,miwa2014modeling,islamaj2011context,he2017insight,uzuner2010extracting}, and general \citep{dethlefs2012optimising,goodman-2002-incremental,frampton2009real,bui2010decision,hsueh2007decisions} domains, there is currently no dataset for extracting (i.e. detecting and classifying) medical decisions in clinical narratives. 
A medical decision is defined as a particular course of clinically relevant actions and/or a statement concerning the assessment of a patient's health as defined in the Decision Identification and Classification Taxonomy for Use in Medicine (DICTUM)~\citep{ofstad2016medical}.

\begin{figure}[t]
    \centering
    \vspace{-20pt}
    \includegraphics[width=\linewidth]{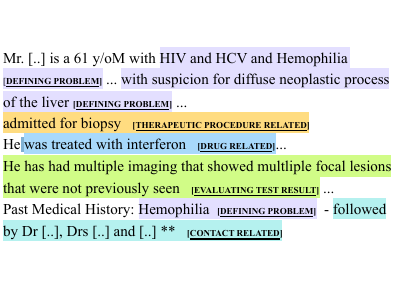}
    \vspace{-40pt}
    \caption{An example excerpt from a de-identified clinical note in MedDec, where text spans are annotated into 10 medical decision categories defined by the Decision Identification and Classification Taxonomy for Use in Medicine (DICTUM)~\citep{ofstad2016medical}. Color-coded texts represent medical decisions and their annotated decision categories are in [\underline{\sc brackets}]. }
    \vspace{-15pt}
    \label{fig:data-sample}
\end{figure}

Automatic extraction of medical decisions from clinical notes has the potential to transform clinical practice. 
It can inform the development of evidence-based decision-making guidelines and stewardship programs, 
identify potential 
deviations from best decision-making practices, and
determine potential risks to patient based on prior medical decisions and their outcomes.
In addition, beyond clinical applications, understanding clinical decision patterns can inform health policy development and refinement, especially when evaluating the impact of particular interventions or policies.



This paper develops the first expert-annotated dataset for medical decision extraction and classification within discharge summaries (MedDec). It is developed using patient data sourced from the Medical Information Mart for Intensive Care
(MIMIC-III) ~\citep{mimic-iii},
which is a publicly available dataset of de-identified clinical data of patients who were treated in intensive care units (ICUs). MedDec contains annotated decisions in 451 discharge summaries, covering more than 54k sentences and containing diverse patient groups based on {\em sex}, {\em race}, and {\em English proficiency}. In addition, 187 out of the 451 discharge summaries were previously classified into eleven phenotypic (main disease) categories through manual annotation by~\citet{gehrmann2018comparing}. We extend the dataset as follows: all medical decisions in the discharge summaries are annotated by domain experts according to the Decision Identification and Classification Taxonomy for Use in Medicine (DICTUM)~\citep{ofstad2016medical}. DICTUM covers ten (10) medical decision categories listed in Table~\ref{tab:decision_cat}; we add the residual category ``None'' for spans of texts that do not contain any medical decision. 
Two expert annotators independently label all text spans of medical decisions in each discharge summary according to the DICTUM guidelines. Disagreements between the annotators are adjudicated by a senior third annotator. Figure~\ref{fig:data-sample} shows an excerpt from a de-identified discharge summary annotated with several categories of medical decisions.

In addition, we introduce the new task of \textit{clinical decision extraction}, which involves identifying and classifying spans of medical decisions within relatively long clinical notes. 
This focused information extraction task contributes to the advancement of bioNLP techniques and has the potential to improve healthcare. 
%
We develop several baselines including span detection and named entity recognition models, and evaluate and analyze their performance on MedDec. In addition, we introduce a framework for medical decision extraction to set a baseline for future research. 

The contributions of this paper are:
\begin{itemize}
\itemsep-2pt
    \item to the best of our knowledge, MedDec is the first expert-annotated dataset for research on medical decision extraction and classification from clinical notes;\looseness-1
    \item we provide a comprehensive analysis of the dataset, including distribution reports for protected variables, including sex, race, and English proficiency; and 
    \item we evaluate existing span detection approaches on MedDec, and develop a baseline model to lay the foundation for future research in this area.  
\end{itemize}

\begin{table*}[t]\footnotesize
    \centering
    \begin{tabular}{lp{7cm} p{4.6cm}}
         \textbf{Decision Category}& \textbf{Description} & \textbf{Examples}  \\
         \toprule
         \textbf{Contact related} & Decision regarding admittance or discharge from hospital, scheduling of control and referral to other parts of the healthcare system & Admit, discharge, follow-up, referral\\
         \midrule
         \textbf{Gathering information} & Decision to obtain information from other sources than patient interview, physical examination and patient chart & Ordering test, consulting colleague, seeking external information \\
         \midrule
         \textbf{Defining problem} & Complex, interpretative assessments that define what the problem is and reflect a medically informed conclusion & Diagnostic conclusion, etiological inference, prognostic judgment \\
         \midrule
         \textbf{Treatment goal} & Decision to set a defined goal for treatment and thereby being more specific than giving advice & Quantitative or qualitative \\
         \midrule
         \textbf{Drug} & Decision to start, refrain from, stop, alter or maintain a drug regimen & Start, stop, alter, maintain, refrain \\
         \midrule
         \textbf{Therapeutic procedure} & Decision to intervene on a medical problem, plan, perform or refrain from therapeutic procedures  & Start, stop, alter, maintain, refrain \\
         \midrule
         \textbf{Evaluating test result} &  Simple, normative assessments of clinical findings and tests & Positive, negative, ambiguous test results \\
         \midrule
         \textbf{Deferment} & Decision to actively delay a decision or rejection to decide on a problem presented by a patient & Transfer responsibility, wait and see, change subject \\
         \midrule
         \textbf{Advice and precaution} & Decision to give the patient advice or precaution, transferring responsibility for action to the patient & Advice or precaution \\
         \midrule
         \textbf{Legal/insurance related} & Medical decision concerning to legal regulations or financial arrangements & Sick leave, drug refund, insurance, disability \\
         \bottomrule
    \end{tabular}
    \caption{Descriptions and high-level examples of medical decisions. The table is re-printed from DICTUM~\citep{ofstad2016medical} with slight modification.}
    \label{tab:decision_cat}
    \vspace{-5pt}
\end{table*}

\begin{table*}[t]
\centering
  \small
    \begin{tabular}{lcccccccccc}    
     \textbf{Decision Type} & \multicolumn{2}{c}{\textbf{Sex}} & \multicolumn{6}{c}{ \textbf{Race}} & \multicolumn{2}{c}{\textbf{Lng. Proficiency}}\\ \toprule
       & {\textbf{Male}} & {\textbf{Female}} & {\textbf{White}} & {\textbf{AA}} & {\textbf{Hispanic}} & {\textbf{Asian}} & 
       {\textbf{NH}} & {\textbf{Other}} & 
       {\textbf{En}} & 
       {\textbf{Non-En}}\\
     
     & (n=259)  & (n=192)  & (n=327) & (n=42) & (n=25) & (n=15) & (n=1) & (n=21) & (n=260) & (n=45)\\ \midrule

 \textbf{Defining Problem} & 39.2 & 38.8 & 39.5 & 37.5 & 38.0 & 36.4 & 30.9 & 38.6 & 38.7 & 39.2\\
\textbf{Drug} & 26.0 & 25.1 & 25.7 & 24.4 & 25.0 & 27.5 & 19.1 & 27.0 & 26.1 & 25.6\\
\textbf{Evaluation} & 12.9 & 13.6 & 12.6 & 16.6 & 13.3 & 12.7 & 25.5 & 12.8 & 13.1 & 13.9\\
\textbf{Therapeutic proc.} & 12.2 & 12.4 & 12.4 & 12.5 & 11.7 & 13.2 & 10.6 & 12.2 & 12.0 & 12.0\\
\textbf{Contact} & 4.9 & 5.2 & 5.0 & 4.6 & 6.0 & 5.4 & 8.5 & 4.3 & 4.8 & 5.1\\
\textbf{Advice} & 3.4 & 3.5 & 3.5 & 3.2 & 4.2 & 3.3 & 0.0 & 3.9 & 3.9 & 3.0\\
\textbf{Gathering info} & 0.8 & 0.9 & 0.8 & 0.7 & 1.2 & 1.3 & 5.3 & 0.9 & 0.9 & 0.6\\
\textbf{Treatment goal} & 0.3 & 0.3 & 0.3 & 0.3 & 0.4 & 0.2 & 0.0 & 0.2 & 0.2 & 0.4\\
\textbf{Deferment} & 0.2 & 0.2 & 0.2 & 0.2 & 0.2 & 0.0 & 0.0 & 0.1 & 0.2 & 0.2\\
\textbf{Legal/Insurance} & 0.0 & 0.0 & 0.0 & 0.0 & 0.0 & 0.0 & 0.0 & 0.0 & 0.0 & 0.0\\

    \midrule
    \textbf{Total Count}	& 33,054 & 24,235 & 41,666 & 5,684 & 3,264 & 1,737 & 94 & 3,078 & 37,026 & 6,295\\
    \bottomrule

    \end{tabular}%
    \caption{Percentage of annotated spans for each decision category across protected variables in MedDec. $n$ is the number of the discharge summaries for each category. The last row shows the total count of decisions per variable. }
    \vspace{-5pt}
    \label{tab:dec_cat}%

\end{table*}%
\begin{table*}
  \small
  \centering
    \begin{tabular}{ p{2.4cm} p{1.1cm} p{0.8cm} p{.8cm} p{.8cm} p{0.8cm} p{0.8cm} p{.8cm} p{.8cm} p{.8cm} p{.8cm} p{.8cm}} 
\textbf{Decision Types} & \multicolumn{10}{c} {\textbf{Phenotypes}}\\ \toprule
 &\textbf{Substance Abuse} & \textbf{Lung} & \textbf{Alcohol Abuse }&  \textbf{Psychi-atric} & \textbf{Obesity} & \textbf{Heart} & \textbf{Cancer}  & \textbf{Chronic Neuro}& \textbf{Depre-ssion} & \textbf{Chronic Pain} &\textbf{None}\\
 & (n=8) &  (n=12) & (n=18) & (n=27) & (n=12) & (n=23) & (n=12) & (n=22) & (n=32) & (n=26) & (n=62)\\
 \toprule

\textbf{Defining Problem} & 38.1 & 36.4 & 40.5 & 38.4 & 37.1 & 38.9 & 38.7 & 40.2 & 36.8 & 36.8 & 38.1\\
\textbf{Drug} & 25.1 & 32.4 & 28.5 & 26.9 & 29.6 & 29.1 & 25.5 & 26.4 & 30.3 & 29.9 & 24.0\\
\textbf{Evaluation} & 16.2 & 10.6 & 11.1 & 14.0 & 12.9 & 11.3 & 10.4 & 14.6 & 13.1 & 13.2 & 14.9\\
\textbf{Therapeutic proc.} & 12.8 & 13.3 & 11.3 & 11.9 & 12.1 & 12.9 & 13.7 & 10.7 & 12.1 & 11.4 & 12.5\\
\textbf{Contact} & 5.5 & 4.4 & 4.1 & 4.9 & 4.8 & 3.9 & 5.2 & 4.3 & 4.4 & 4.7 & 5.7\\
\textbf{Advice} & 1.1 & 1.5 & 2.9 & 2.7 & 2.5 & 2.7 & 4.7 & 3.0 & 2.4 & 2.8 & 3.6\\
\textbf{Gathering info} & 0.6 & 0.9 & 1.0 & 0.9 & 0.3 & 0.8 & 1.2 & 0.5 & 0.6 & 0.9 & 0.7\\
\textbf{Treatment goal} & 0.4 & 0.3 & 0.2 & 0.1 & 0.3 & 0.3 & 0.3 & 0.1 & 0.1 & 0.1 & 0.3\\
\textbf{Deferment} & 0.2 & 0.2 & 0.3 & 0.2 & 0.4 & 0.1 & 0.2 & 0.2 & 0.2 & 0.2 & 0.2\\
\textbf{Legal/Insurance} & 0.0 & 0.0 & 0.0 & 0.0 & 0.0 & 0.0 & 0.0 & 0.0 & 0.0 & 0.0 & 0.0\\

 
\midrule
\textbf{Total Count} & 2,062 & 4,319 & 4,464 & 8,726 & 2,957 & 7,126 & 2,271 & 8,301 & 10,289 & 8,639 & 16,790\\
\bottomrule

    \end{tabular}%
    \caption{Percentage of annotated spans for each decision category across different phenotypes. $n$ is the
number of the discharge summaries for each category. The last row shows the total count of decisions for each phenotype.}
\vspace{-10pt}
    \label{tab:phenotype}%

\end{table*}%
     



\section{MedDec}


\subsection{Taxonomy of Medical Decisions}
Table~\ref{tab:decision_cat} provides descriptions of different types of medical decisions in clinical notes, adapted from DICTUM~\citep{ofstad2016medical}. 
The ``Contact related" category involves decisions related to patient admissions, discharges, follow-ups, and referrals within the healthcare system. 
``Gathering information" decisions pertain to acquiring data from sources other than patient interviews or charts, such as ordering tests.  
``Defining problem" decisions involve complex assessments that define medical issues, including diagnoses, etiological inferences, and prognostic judgments. 
``Treatment goal" decisions specify treatment objectives beyond general advice. 
``Drug" decisions pertain to initiation, alteration, cessation, or maintenance of drug regimens. 
``Therapeutic procedure" decisions involve interventions or therapeutic procedure management. 
``Evaluating test result" decisions are those that evaluate clinical findings and test outcomes. 
``Deferment" decisions delay or reject medical decision-making, often due to insufficient information or the need to await test results. 
``Advice and precaution" decisions involve providing advice or precautions to patients and transferring responsibility for actions to them. 
Finally, ``Legal/insurance-related" decisions deal with medical matters related to legal regulations or financial arrangements. 

These categories provide a comprehensive grouping of medical decisions in clinical notes. They can be used for systematic classification and structured analysis of medical decisions, and for understanding the complex processes involved in clinical decision making.

\subsection{Data Collection}
MedDec is created using patient data sourced from the MIMIC-III~\citep{mimic-iii}. It contains annotated decisions in 451 discharge summaries, representing diverse patient groups based on sex, race, and English proficiency. All medical decisions in each discharge summary are annotated according to the 10 medical decision categories introduced in DICTUM~\citep{ofstad2016medical}. 

The token-level inter-annotator agreement, measured by Cohen's Kappa between the first two annotators is substantial, $k = 0.74$, indicating that it is fairly easy for domain experts to identify medical decisions in discharge summaries. A similar agreement level was reported in DICTUM~\citep{ofstad2016medical}.  We note that token-level agreement provides a lower bound for true inter-annotator agreement as it may sometimes underestimate agreement. This occurs, for instance, when minor variations such as the inclusion or exclusion of less relevant tokens (e.g. stopwords) at the start or end of decision spans are considered as disagreements. 

\subsection{MedDec Novelty}
The novelty of MedDec is in its focused annotation of medical decisions based on an expert-verified and comprehensive taxonomy of medical decisions, its diversity across sex, race, and language proficiency patient groups and phenotypes (diseases), and its potential to drive advancements in both bioNLP research and clinical decision-making. To our knowledge, MedDec is the first dataset specifically developed for extracting medical decisions in clinical notes. This diversity in MedDec enables investigations on potential disparities in medical decisions across the above-mentioned protected variables, which can provide insights for addressing healthcare inequities. 
These features make MedDec an asset in bioNLP.

\subsection{MedDec Statistics}
Table~\ref{tab:dec_cat} reports the percentage of decision spans for each decision category and each protected variable in MedDec. The total counts of decision spans are reported in the last row. Medical decisions related to defining problems, drugs, evaluation, and therapeutic procedures are categories with the highest prevalence, while legal, deferment, treatment goal, gathering information, advice, and contact have considerably lower prevalence. In MedDec, 42.6\% of summaries are related to Female patients, 75.9\% belong to white patients (of patients with known race), 9.7\% to African American, and 85.2\% to patients (with known language proficiency) identified as proficient in the English language. 

In addition, Table~\ref{tab:phenotype} shows the distribution of patients across phenotypic (disease) categories. Patients with psychiatric disorders (including schizophrenia, bipolar, and anxiety disorders), depression, chronic neurologic dystrophies, and chronic pain are more prominently represented in MedDec. Conversely, patients associated with substance abuse, lung conditions, cancer, and obesity are less prevalent in the dataset.

The discharge summaries in MedDec contain 1.4M tokens, with 879K tokens forming part of a span, while 37K tokens belong to more than one span (accounting for 4.2\% of labeled tokens). However, the majority of overlaps are minor, where a token marks the end of a span and the start of another.\looseness-1

\section{Learning to Extract Medical Decisions}
\begin{figure}[t]
    \centering
    \includegraphics[width=1\linewidth]{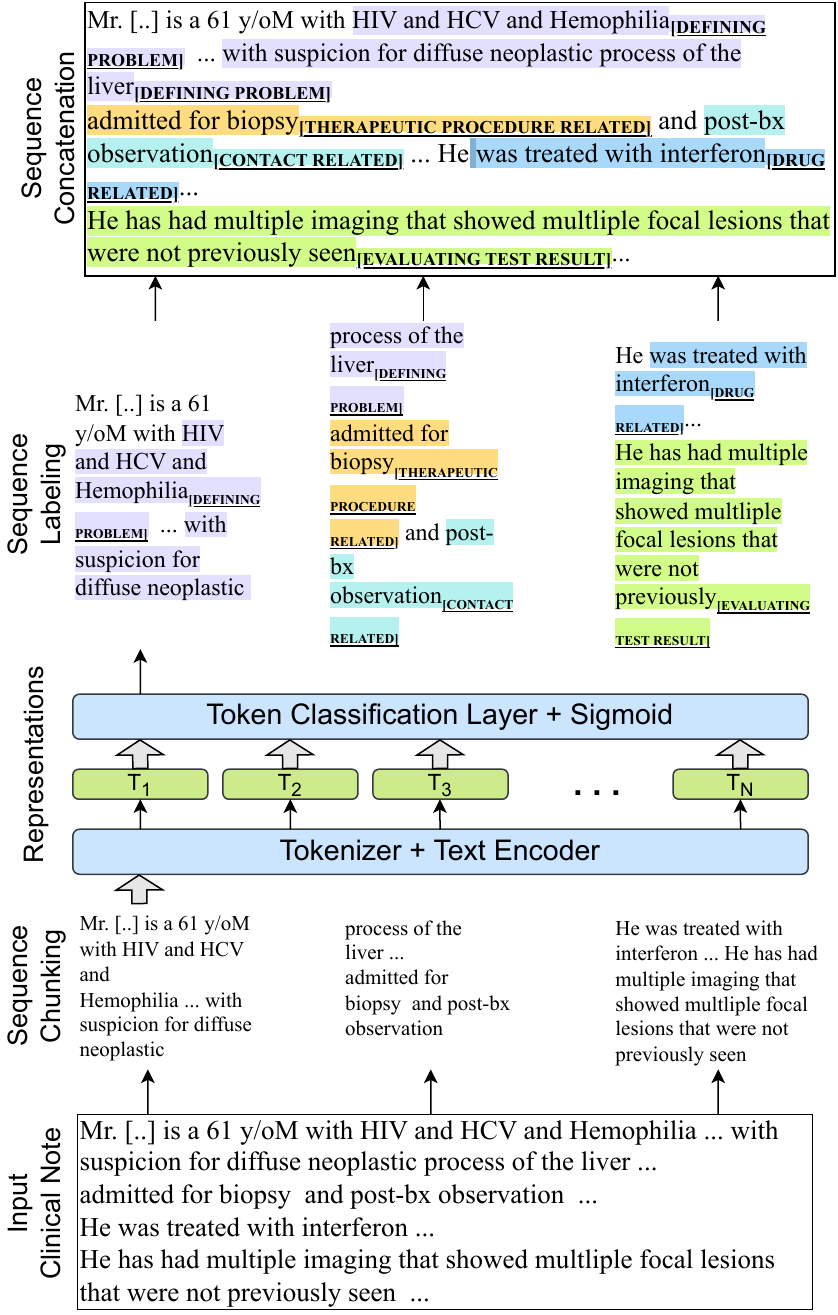}
    \caption{Architecture of the proposed framework for medical span detection. The framework is a multi-class sequence labeling approach that fine-tunes a pre-trained transformer network for span detection.}
    \vspace{-5pt}
    \label{fig:arch}
\end{figure}

We present an overview of our approach and describe its two key components in subsequent sections.
Figure~\ref{fig:arch} shows our approach for extracting and classifying medical decisions. First, a {\em long} note is chunked into segments of acceptable length to the model. Each segment is fed into the model to generate hidden representations and token classification probabilities (this step can be batched for efficiency). 
Then, the resulting labeled text sequences are concatenated to obtain classification results for the entire clinical note. 
Finally, we post-process the results to convert them into spans, defined by a start position, end position, and category.

\subsection{Sequence Labeling Framework}
We develop a multi-class sequence labeling approach that fine-tunes a pre-trained model for span detection.
The data consists of a sequence of $n$ tokens $t = \{t_1,\dots,t_n\}$ and token labels $y = \{y_1,\dots,y_n\}$, and each label indicates a set of $k+1$ categories indicating $k$ decision types and the {\tt none} category, $ y_i \in \{C_1, .., C_k, O\}$. Practically, the labels follow the BIO (beginning-inside-outside) token labeling scheme~\citep{ramshaw1999text}.

We use a pre-trained bidirectional transformer encoder $f$ to encode the tokens and generate $d$-dimensional hidden states. Formally, the latent representation of the $i$-th token is computed as:
\begin{equation}
    \mathbf{h}_i = f(t)_i,
\end{equation} 
where $\mathbf{h}_i \in \mathbb{R}^d$.
We then employ a fully-connected layer $g(.)$ on top of the hidden representation that maps each hidden state to obtain the logits across all classes of decision categories:
\begin{equation}
    z_{ik} = g(h_i)_k.
\end{equation}
To realize multi-class classification, the logits are fed into a softmax function, where the class with the maximum predicted probability is considered the predicted label. This approach does not take into account the categorical similarity of neighboring tokens. Previous studies indicate that token classification remains competitive as the most effective span detection and named entity recognition method~\citep{aarsenspanmarker, 
jurkiewicz-etal-2020-applicaai, chhablani-etal-2021-nlrg, gu-etal-2022-empirical}.


There are several alternative methods for span detection, such as detecting the start and end positions, classifying a pair of tokens to check if they constitute a span's boundary, detecting each span category using a separate query process~\citep{devlin-etal-2019-bert,shen-etal-2022-parallel}, and conditional random fields~\citep{panchendrarajan-amaresan-2018-bidirectional} that compute the maximum probable span assignment from logits.\footnote{Our experiments show that CRF does not improve performance; so, we do not include it in our architecture.}

\subsection{Sequence Chunking}
Clinical notes are typically thousands of tokens long, and transformer models are computationally restricted and can typically process a maximum of 512 tokens at once.
To overcome this challenge, we develop a data sampling function that samples segments of 512 tokens or fewer from random starting points at each training iteration. Therefore, a unique set of text segments is seen at each iteration. This sampling method acts as a data augmentation method by sampling different segments of the same clinical note with different start and end positions, different sentence compositions, and different lengths. 
At inference time, the input is chunked into segments of 512 tokens with no overlap, each segment is tagged, and the results are concatenated.

\begin{table*}[t]\small
    \centering
    \begin{tabular}{l lc lllllllll}
    \textbf{Model} &  \textbf{Token Level} &\textbf{Span Level}  & \textbf{CR} & \textbf{GI} & \textbf{DP} & \textbf{TG} & \textbf{Dr} & \textbf{TP} & \textbf{ETR} & \textbf{De} & \textbf{A\&P} \\
     &  (Accuracy) &(F1)  & (F1)  & (F1)  & (F1)  & (F1)  & (F1)  & (F1)  & (F1)  & (F1)  &(F1)  \\
    \toprule
         \textbf{ELECTRA} & 78.2 & 34.7 & \textbf{19.9} & 0.0 & 37.9 & 0.0 & 47.4 & 25.2 & 19.7 & \textbf{15.4} & 35.1 \\
         \textbf{BioClinicalBERT} & 77.8 & 34.5 & 15.9 & 4.2 & 38.9 & \textbf{11.8} & 46.4 & 27.0 & 19.4 & 0.0 & 33.8 \\
         \textbf{RoBERTa} & \textbf{79.9} & \textbf{34.8} & 19.3 & \textbf{5.1} & 37.3 & 6.1 & 44.7 & 27.9 & 23.4 & 12.5 & \textbf{42.6}\\
         \textbf{DeBERTa v3} & 77.4 & 31.9 & 15.2 & 2.2 & 32.7 & 7.4 & 46.8 & 24.6 & 18.5 & 0.0 & 28.0\\
         \textbf{ALBERT v2} & 74.6 & 27.8 & 10.9 & 4.1 & 33.0 & 0.0 & 38.8 & 16.6 & 15.2 & 0.0 & 12.0\\
         \midrule
         \textbf{BINDER}         &  71.2& 30.3& 17.4& 2.5& \textbf{59.6}& 1.0& \textbf{50.9}& \textbf{36.8}& \textbf{34.0}& 0.9& 10.2\\
         \textbf{PIQN}           &  69.5& 28.9& 16.9& 2.4& 57.6& 1.0& 48.9& 33.8& 32.7& 0.9& 9.1
\\
         \textbf{DyLex}          &  67.7& 27.8& 17.4& 2.4& 57.1& 0.9& 46.0& 31.7& 30.1& 0.8& 0.9\\
         \textbf{Instance-based} &  66.2& 27.0& 16.1& 2.5& 56.7& 0.9& 44.3& 31.8& 28.7& 0.8& 8.4\\
         \bottomrule
    \end{tabular}
    \caption{
    Span detection performance of different models on MedDec. Span level evaluates the exact match at the span level, while token level evaluates the prediction of decision categories for individual tokens in inputs.  \em{Columns 4-11 show the performance on each decision category}, abbreviated according to the order in Table~\ref{tab:decision_cat}}
    \label{tab:main_res}
\end{table*}

\section{Experiments}

\subsection{Experimental Setup}

\paragraph{Models} We evaluate the following models:
\begin{itemize}
    \itemsep-2pt
    \itemindent-10pt
\item{\bf Binder}~\citep{zhang2023optimizing}: employs encoders for tokens and token types, optimizes a contrastive objective, with a dynamic threshold loss for negative sampling.

\item{\bf PIQN}~\citep{shen-etal-2022-parallel}:
uses NER pointer mechanism~\citep{yang-tu-2022-bottom} for span boundary detection and an entity classifier for classification. A dynamic label assignment objective is proposed to assign gold labels to instance queries. It dynamically learns query semantics for instance queries and extracts all types of entities simultaneously.

\item{\bf DyLex}~\citep{wang-etal-2021-dylex}: a sequence labeling-based approach that incorporates lexical knowledge with an efficient matching algorithm to generate word-agnostic tag embeddings for NER.

\item {\bf Instance-based NER}~\citep{ouchi-etal-2020-instance}: 
formulates NER as instance-based learning, where model 
assigns labels based on a nearest-neighbor approach. 
\end{itemize}

\noindent The following BERT-based models employ the token classification approach described in \citet{devlin-etal-2019-bert}. All experiments use the base-size version of the models.
\begin{itemize}
    \itemsep-2pt
    \itemindent-10pt
\item {\bf DeBERTa v3}~\citep{he2022debertav3}: 
uses advanced training strategies, primarily disentangled attention and mask decoder.

\item {\bf ALBERT}~\citep{lan2019albert}:
implements shared weights across layers, leading to a greatly reduced memory footprint.

\item {\bf ELECTRA}~\citep{clark2019electra}: a bidirectional encoder employing a new pre-training objective. It learns to discriminate between real and fake (but plausible) input tokens.

\item {\bf RoBERTa}~\citep{liu2019roberta}: a BERT-based model with an improved training objective, hyperparameters, and increased data.

\item {\bf BioClinicalBERT}~\citep{alsentzer-etal-2019-publicly}: a BERT-based model pre-trained on PubMed abstracts and MIMIC-III clinical notes. 

\end{itemize}

\paragraph{Evaluation}
We use the standard evaluation metrics for NER, span exact match, and token accuracy.\footnote{\url{https://huggingface.co/spaces/evaluate-metric/seqeval}} The correctness of a span is determined by an exact match (both boundaries and category). We report results in terms of micro-F1 score. Token-level accuracy is a more flexible metric, allowing partial overlap with the true spans.

\paragraph{Difficulty Score}
We use span length and number of UMLS medical concepts to divide medical decisions into into three difficulty levels, namely Easy, Medium, and Hard.
Example of a span with a single medical concept and low cognitive load: ``you will continue taking two \underline{antibiotics}.''
Example of a span with a high number of medical concepts (underlined) and high cognitive load:
``\underline{CT abdomen with intravenous contrast}: The \underline{heart size} is at the upper limits of \underline{normal}. \underline{Dense coronary calcifications} are identified. In the \underline{lung} bases, there is \underline{bibasilar atelectasis}. There are also \underline{chronic pleural inflammatory changes} including \underline{fat deposition} and \underline{fibrotic changes}, \underline{left} greater than \underline{right}. \underline{Bilateral small pleural effusions} are also identified, \underline{right} greater than \underline{left}. No \underline{focal pulmonary nodules} or \underline{opacities} are identified in the \underline{lung} bases.''
These difficulty scores provide insights into the
difficulty of learning medical decisions and can also inform curriculum discovery~\citep{elgaar-amiri-2023-hucurl}.

\begin{figure}
    \centering
    \includegraphics[width=.8\linewidth]{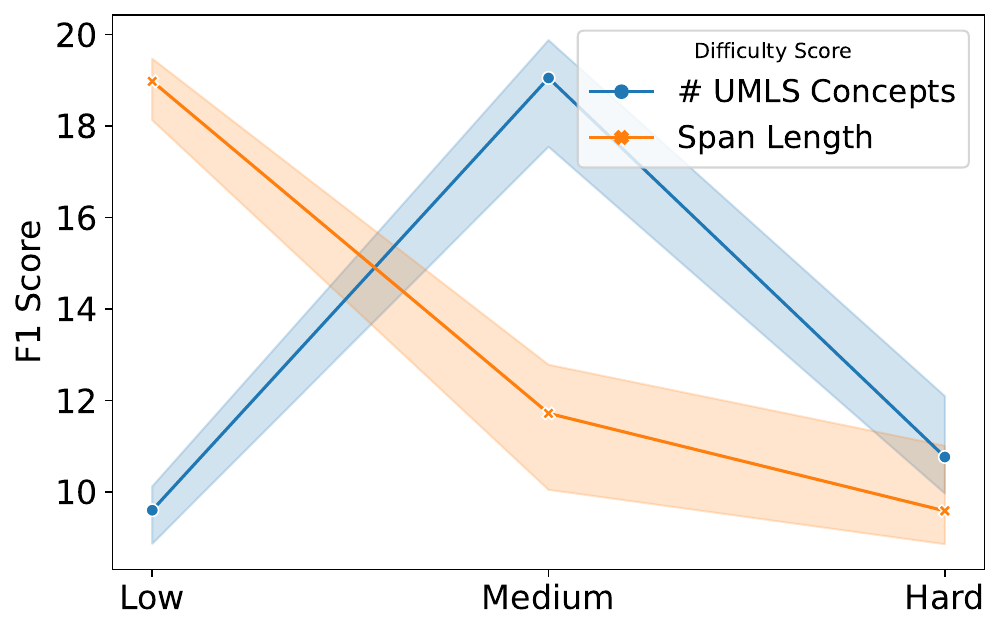}
    \vspace{-5pt}
    \caption{Span detection F1 score on spans with increasing difficulty for two difficulty scores. The shaded area is the 95\% confidence interval for three models: ELECTRA, RoBERTa, and BioClinical-BERT.}
    \label{fig:loss_len_diff}
    \vspace{-10pt}
\end{figure}

\begin{figure*}
    \centering
    \includegraphics[width=.9\linewidth]{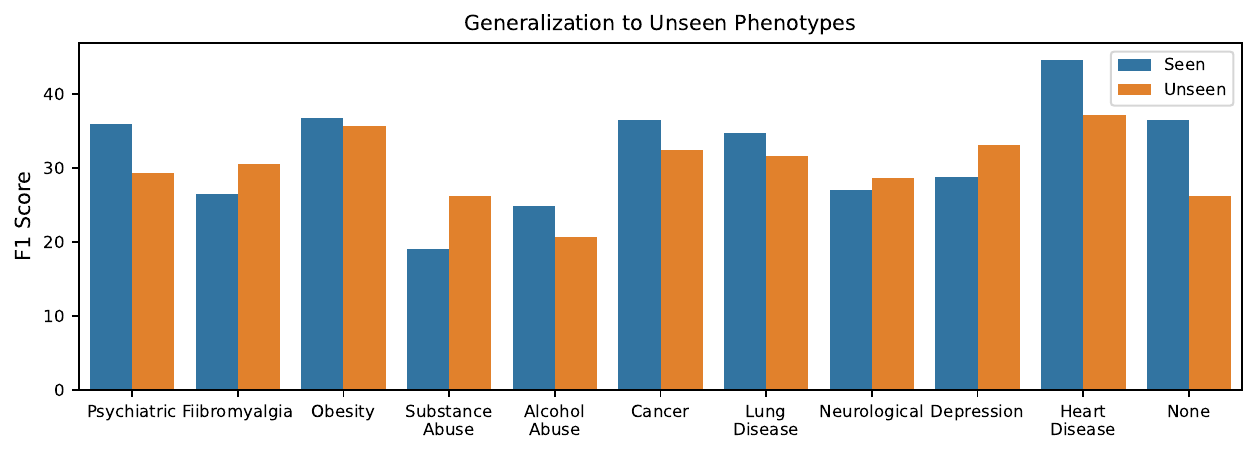}
    \vspace{-10pt}
    \caption{F1 score performance of span detection at phenotype level. The orange bars show the generalizability performance of the model when the phenotype is {\em unseen} during training.}
    \label{fig:phenotypes}
\end{figure*}

 
\subsection{Main Results}
We compare recent span detection and classification models using our training framework.\footnote{We note that the performance of current models without our training framework is significantly lower.}
Table~\ref{tab:main_res} shows the results where we observe that RoBERTa achieves the best performance with a 34.8 F1 score, followed by ELECTRA and BioClinicalBERT.

Span and token accuracy are not perfectly correlated. Although ALBERT performs lower than BINDER in span exact-match, it achieves a higher token accuracy, meaning that it is more effective in partial span detection. BINDER, PIQN, and DyLex achieve higher span-level but lower token-level accuracy than ALBERT, stemming from their design that emphasizes span exact matching.

Our approach focuses on training a model to reliably label segments in clinical notes, irrespective of their boundaries. This strategy results in improved span boundary prediction at inference time.



\subsubsection{Span Complexity}
We suggest two factors that offer insights into the difficulty of learning spans of medical decisions: 
(a): the number of medical concepts in each span according to the UMLS ontology, and 
(b) the length of spans as a heuristic metric; the longer spans may contain more diverse semantic content and can involve more complex sentence structures, which requires the model to maintain more contextual information. We divide the spans into three groups based on their complexity: low, medium, and high.

The results show significant performance disparity across three complexity levels and two metrics. In particular, performance varies considerably based on span length, where predictions are most accurate on shorter spans and least accurate on longer ones. The performance reaches up to 17.7 on easy samples and as low as 10.0 on hard samples. However, we do not observe such a decreasing trend in performance using the number of UMLS concepts, see Figure~\ref{fig:loss_len_diff}. Span detection performance drops when there is either a very low or very high number of medical concepts in a span. A span with a low number of medical concepts is likely ambiguous and hard to interpret and classify, while a span with a large number of medical concepts is complex and hard to understand. Therefore, sentence length is a better measure of sample complexity, perhaps due to the use of broader contextual information. 

We note that the difficulty analysis is the average of the following three models, ELECTRA~\citep{clark2019electra}, RoBERTa~\citep{liu2019roberta}, and BioClinicalBERT~\citep{alsentzer-etal-2019-publicly}. Comparing model performance across varying levels of sample complexity provides a better understanding of each model's strengths and weaknesses.


\subsubsection{Insights Across Patient Phenotypes}
Figure~\ref{fig:phenotypes} shows a significant disparity in the performance of span detection at the phenotype level, with a standard deviation of 7.5. The highest F1 score is achieved for ``Heart Disease'' (44.7), while the lowest is for ``Substance Abuse'' (19.1). Across 11 phenotypes, performance exceeds the average performance (from Table~\ref{tab:main_res}) for six phenotypes but falls under the average for the remaining five phenotypes:
``Depression,'' ``Neurological Dystrophies,'' ``Substance Abuse,'' ``Fibromyalgia'' and ``Alcohol Abuse''. 
The variability in the performance of decision extraction across different patient groups identified by different phenotypes highlights the challenges associated with the practical use of the system.
We attribute this variance in performance to two potential factors: 
(a): MedDec contains more training data for the high-performing phenotypes and less training data for the low-performing phenotypes, 
(b): the intrinsic characteristics of phenotypes affect their difficulty in learning. For example, ``Heart Disease'' is identifiable through clear clinical markers like abnormal ECG findings or chest pain. Conversely, ``Fibromyalgia'' is a more complex condition due to its complex nature and subjective symptoms like widespread
pain and fatigue. The subjective nature of these symptoms and their overlap with other conditions make it challenging to precisely classify ``Fibromyalgia" cases.\looseness-1

\subsubsection{Generalizability to Unseen Phenotypes}
Results in Figure~\ref{fig:phenotypes} show a performance drop for 7/11 phenotypes when they are not present in the training data. Conversely, Substance Abuse, Fibromyalgia, Depression, and Neurological do not suffer from being unseen, as perhaps they are sufficiently informed by transferred knowledge from available phenotypes in the training data. Heart disease and Psychiatric disorders are among the most affected phenotypes, perhaps due to specific domain knowledge related to their decisions that differentiate them from others. 

\subsubsection{IFT for Span Extraction}

\begin{table}[t]
\small
    \centering
    \begin{tabular}{c c c}
        \textbf{Method} & \textbf{F1 (exact match)} & \textbf{F1 (fuzzy match)} \\
        \toprule
        \textbf{Zero-shot} & 3.8 & 10.4 \\
        \textbf{One-shot} & 4.8 & 17.9 \\
        \bottomrule
    \end{tabular}
    \caption{Preliminary analysis of span extraction performance for a prompted LLM in terms of F1 scores of exact and fuzzy match on 10 discharge summaries.}
    \label{tab:gen_res}
    \vspace{-10pt}
\end{table}

Large language models (LLMs) have shown effectiveness in performing a wide variety of tasks through instruction-tuning (IFT)~\citep{zhao2023survey,li-etal-2024-llamacare-instruction,tran2024bioinstruct}. In tasks such as medical question answering, recent works have shown that LLMs show comparable 
performance to extensively fine-tuned models~\citep{nori2023capabilities,singhal2023towards,thirunavukarasu2023large} using domain-specific prompting methods, such as the retrieval of relevant medical queries to serve as demonstrations~\citep{nori2023can}.

We evaluate \texttt{LLama-3-8B-Instruct}~\citep{llama3modelcard} on 10 discharge summaries. We prompt the LLM to extract decision spans for each decision category separately, prompting it ten times for each clinical note.
We experiment with the zero-shot and one-shot settings both using the following prompt:
\begin{verbatim}
[[[System]]]
Extract all sub-strings from the 
following clinical note that contains
medical decisions within 
the specified category.
Print each sub-string in a new line.
If no such sub-string exists, output "None".
[Clinical Note]: {Discharge summary here}

# IF: one-shot setting
[[[User]]]
[Category]: {Decision category here}

[[[Assistant]]]
{Demonstrations}
# End IF

[[[User]]]
[Category]: {Decision category here}

[[[Assistant]]]
{Response}
\end{verbatim}

In the one-shot setting, we present as demonstrations all decisions of a single category other than the one being asked for. The demonstration category is selected as the category with the most number of decisions in the clinical note.
The results are shown in Table~\ref{tab:gen_res}. The LLM returns the extracted spans without token-level annotations, therefore it is not possible to calculate token-level accuracy. We compute the performance of span exact match and fuzzy match F1-score. The span fuzzy match is a substitute for token-level accuracy used in Table~\ref{tab:main_res}, as the span may be partially detected by the LLM but not be accounted for by the exact match score. To compute the fuzzy span match, we check if either of the extracted span and the true span are a substring of the other, and that they differ by no more than 10 words.
These preliminary results show that decision extraction might be challenging for LLMs compared to fine-tuned models.

The low performance of the IFT models can be attributed to the lower efficacy of LLMs in processing long contexts~\citep{an2023eval}. Moreover, the output of LLMs is in free form, which can result in correct responses that do not precisely match the content of medical decisions in notes. For example, a documented decision can be ``the patient has high blood pressure,'' whereas the generated decision can be ``the patient is experiencing elevated blood pressure.'' While semantically correct and relevant, such responses make accurate evaluation of LLM responses a challenging task. 


\section{Related Work}

\subsection{BioNLP Datasets}
\citet{nye-etal-2018-corpus} developed a dataset of 5K abstracts annotated with \{\textit{population, interventions, compared, outcomes}\} (PICO), to inform personalize patient care.
\citet{lehman-etal-2019-inferring} developed a dataset of intervention, comparator, and outcome labels of more than 10K randomized controlled trial articles.
\citet{patel-etal-2018-annotation} developed the clinical entity recognition (CER) corpus, which consists of 5.1K clinical notes annotated by experts 
with entities such as 
anatomical structures, body functions, 
lab devices and medical problems.
, and findings. 
The extracted concepts correspond to a selected group of UMLS semantic types.
CLIP~\citep{mullenbach-etal-2021-clip} is a dataset of 718 discharge summaries from MIMIC-III, annotated with seven types of action items: \{\textit{Appointment, Lab, Procedure, Medication, Imaging, Patient Instructions, Other}\} at sentence level. 
covering more than 107K sentences. 
%
\citet{structured_understanding} introduced a dataset of 579 admission and progress notes from MIMIC-III, annotated with diseases, assessments, and categories of action items. 
%
PHEE~\citep{sun-etal-2022-phee} consists of 5K events from case reports and literature, annotated for pharmacovigilance \{\textit{Subject, Drug, Effect}\} for drug safety monitoring. 
Recently, \citet{cheng-etal-2023-mdace} developed MDACE, a dataset of clinical notes annotated with ICD codes and their rationales for computer-assisted coding.
%

\subsection{NER and Span Detection}
Existing NER and span detection approaches can be divided into sequence labeling-based (tagging) approaches and span-based approaches.
Sequence labelling approaches~\citep{aarsenspanmarker,sang2003introduction,gu-etal-2022-empirical}
classify every token in the sequence to their corresponding class(es). This formulation is 
challenged by nested entities. 
Span-based approaches~\citep{sohrab-miwa-2018-deep,luan-etal-2019-general,zheng-etal-2019-boundary,tan2020boundary,shen2021}, however, identify and classify spans. 
First, the spans are either extracted through enumeration or boundary identification and then classified to their corresponding classes. 

~\citet{du-etal-2019-learning} developed the relational span-attribute tagging (R-SAT) model for extracting clinical entities, their properties, and relations. It employed a method similar to ours, however, the tasks are different as we tackle medical decisions.
\citet{ouchi-etal-2020-instance} formulated the NER task as an instance-based learning task, where the NER model was trained to learn the similarity between spans of the same class. 
%
%
DyLex~\citep{wang-etal-2021-dylex} 
retrieved lexicon knowledge for input sequence, applied a denoising module to remove noisy matches, and encoded and fused the lexicon knowledge into the sequence embeddings with column-wise attention for NER.
%

\citet{abaho-etal-2021-detect} jointly detected and classified spans of health outcomes in clinical notes. Most prior methods decoupled the detection and classification phases. 
%
%
Sent2Span~\citep{liu-etal-2021-sent2span-span} was developed for the extraction of PICO information from clinical trial reports.
It is designed to work with non-expert sentence-level annotations on the presence of PICO information, without the need of expert span-level annotations and is able to achieve higher recall than comparable methods. 
PIQN~\citep{shen-etal-2022-parallel} 
developed entity pointer for localization and entity classifier for classification. A dynamic label assignment objective was proposed to extract different types of entities simultaneously. 

Recently, \citet{zhang2023optimizing} proposed Binder, which optimized two encoders for 
NER, one for tokens and one for token types. 
For each span associated with a class, Binder sampled negative spans based on their loss, and optimized model parameters using a contrastive learning objective. 
Mirror~\citep{zhu-etal-2023-mirror} was an information extraction framework based on graph decoding, where entities were nodes in the graph and the relations of interest were edges. Mirror allowed for extracting all entities and relations in a single step. 
DICE~\citep{ma-etal-2023-dice} adapted sequence-to-sequence models for structured event extraction from clinical text, using PubMed documents in MACCROBAT dataset~\citep{caufield2019comprehensive}.
\citet{raza2023constructing} introduced a model consisting of BioBERT~\citep{biobert}, and Bi-LSTM~\citep{bilstm} and Conditional Random Field (CRF)~\citep{crf} layers to extract clinical (diseases, conditions, symptoms, and drugs) and non-clinical (social factors) entities from clinical notes. 

\section{Conclusion}
We developed MedDec for the extraction and classification of ten types of documented medical decisions in discharge summaries of eleven different phenotypes (diseases). We demonstrate several baseline models to tackle this task. Through extensive experiments and analysis, we find that the task is challenging, the performance of the best-performing model significantly varies across phenotypes and the spectrum of sample complexity. The dataset can be useful in studying population statistics, biases in medical treatment, analysis of medical decisions for different phenotypes, and understanding medical decision-making processes. 


\section*{Limitations}
The present work has several limitations, which form the basis of our future work: 
the distribution of the ten classes of medical decisions within the dataset is considerably imbalanced. Table~\ref{tab:dec_cat} highlights these data imbalances across protected variables. Class imbalance may lead to biases in the model training process and affect the model's ability to accurately predict less represented classes. Addressing these data imbalances can prevent computational models from learning and perpetuating such biases in the data. In addition, we note that, these imbalances reflect the challenges of working with real-world data and can inform future research in healthcare equity and the development of systems that perform well across all patient groups.
The models have been applied to the notes in MIMIC-III dataset, and the extent of their generalizability to other datasets has not been evaluated. While our best classifier is performant, it may fail to identify and classify certain medical decisions, such as those pertinent to deferment. This limitation could be partly due to the effect of longer decisions, which can inversely affect the model's performance due to the potentially higher linguistic complexity of longer texts (see relevant results in Figure~\ref{fig:loss_len_diff}).
%
Finally, while discharge summaries contain rich information about patient care, it's important to acknowledge their limitations in fully capturing the breadth of medical decisions made during a patient's hospital stay. Nevertheless, common medical decision-making patterns and clinical reasoning processes are expected to make models trained on these summaries generalize to other types of clinical documents.



\section*{Ethic and Broader Impact Statements}
This project adheres to ethical considerations and safeguards to ensure the responsible and ethical handling of medical data and its implications. All results
have been presented in aggregate, and 
we have made and will make every effort to protect human subject information and minimize the potential risk of loss of patient privacy and confidentiality (all
authors with access to the data have successfully
completed a training program in the protection
of human subjects and privacy protection). In addition, our work is transformational in nature, and its
broader impacts are first and foremost the potential
to improve the well-being of individual patients
in the society
and support clinicians in their
medical decision-making efforts.

\bibliography{anthology,anthology_part2,custom}
\onecolumn

\appendix
\section{Examples of Model Predictions}
\begin{table}[t]
\small
    \centering
    \begin{tabular}{p{0.5\textwidth}>{\raggedright\arraybackslash}p{0.5\linewidth}}
        \textbf{Text} & \textbf{Context}\\
        \toprule
        \dots during that admission, \falsepred{he coded for}\truepred{pulseless vt} and was transferred &  The first three words of the "Defining Problem" span are not detected.\\
        \hline
        chief complaint: \falsepred{shortness of breath}. & Model failed to extract "Defining Problem" decision. \\
        \hline
        \dots were orally administered. \truepred{The patient demonstrated piecemeal behavior by dividing up boluses into multiple swallows regardless of size of consistency of the bolus.}\falsepred{There was subsequent premature spillover into the valleculae.} & The first sentence is correctly classified as "Evaluating test results". The second sentence is incorrectly classified as "Defining problem" instead of "Evaluating test results". \\
        \hline
 \truepred{Coronary angiography in \dots 90\% stenosis after d1.}\falsepred{The}\truepred{lcx was totally occluded \dots via left collaterals.}\falsepred{The}\truepred{rca had a 90\% proximal lesion.} & Three separate "Evaluating test results" decisions are detected as one. The decision boundary is incorrectly classified. \\
        \hline
        \truepred{He was transferred to [..]}and\falsepred{neurosurgery was consulted.} & The first segment is correctly classified as "Contact related", the second segment is incorrectly classified as "Contact related" instead of "Gathering additional information". \\
        
    \end{tabular}
    \caption{Examples where the model fails to extract medical decisions.}
    \label{tab:examples}
\end{table}

Table~\ref{tab:examples} shows examples where the model partially or fully fails to capture the underlying medical decision from the clinical note.

\end{document}